\documentclass[sigconf]{acmart}

\AtBeginDocument{%
  \providecommand\BibTeX{{%
    \normalfont B\kern-0.5em{\scshape i\kern-0.25em b}\kern-0.8em\TeX}}}

\setcopyright{rightsretained}
\acmConference[EARS '20]{ACM SIGIR International Workshop on Explainable Recommendation and Search}{July 30, 2020}{Virtual Event, China}
\acmBooktitle{EARS '20: ACM SIGIR International Workshop on Explainable Recommendation and Search,
July 30, 2020, Virtual Event, China}
\acmISBN{978-1-4503-8016-4/20/07}
\acmPrice{15.00}
\copyrightyear{2020}
\acmYear{2020}
\acmDOI{}

\acmConference[SIGIR '20]{SIGIR '20: ACM Conference on Research and Development in Information Retrieval}{July 25-30, 2020}{Xian, China}

\begin{document}

\title{Generating medical screening questionnaires through analysis of social media data}

\author{Ortal Ashkenazi}
\affiliation{%
 \institution{Technion}
 \country{Israel}}
\email{ortasa@gmail.com}

\author{Elad Yom-Tov}
\affiliation{%
 \institution{Technion and Microsoft Research}
 \country{Israel}}
\email{eladyt@microsoft.com}

\author{Liron Vardi David}
\affiliation{
 \institution{Hillel Yaffe Medical Center} 
 \country{Israel}}
\email{Lironda89@gmail.com}

\newcommand{\elad}[1]{\textcolor{red}{#1}}
\newcommand{\ortal}[1]{\textcolor{blue}{#1}}

\begin{abstract}

Screening questionnaires are used in medicine as a diagnostic aid. Creating them is a long and expensive process, which could potentially be improved through analysis of social media posts related to symptoms and behaviors prior to diagnosis. 
Here we show a preliminary investigation into the feasibility of generating screening questionnaires for a given medical condition from social media postings. The method first identifies a cohort of relevant users through their posts in dedicated patient groups and a control group of users who reported similar symptoms but did not report being diagnosed with the condition of interest. Posts made prior to diagnosis are used to generate decision rules to differentiate between the different groups, by clustering symptoms mentioned by these users and training a decision tree to differentiate between the two groups. We validate the generated rules by correlating them with scores given by medical doctors to matching hypothetical cases. 
We demonstrate the proposed method by creating questionnaires for three conditions (endometriosis, lupus, and gout), using the data of several hundreds of users from Reddit. These questionnaires were then validated by medical doctors. The average Pearson’s correlation between the latter's scores and the decision rules were 0.58 (endometriosis), 0.40 (lupus) and 0.27 (gout). 
Our results suggest that the process of questionnaire generation can be, at least partly, automated. These questionnaires are advantageous in that they are based on real-world experience, but are currently lacking in their ability to capture the context, duration, and timing of symptoms.

\end{abstract}



\maketitle

\section{Introduction}

The internet is a primary source for medical information for people who require medical information. Pew surveys indicated that 80\% of U.S. internet users reported looking for health information on the internet \cite{fox2011social}. Though search engines are often the first source of information \cite{fox2011social}, social media, especially when anonymous \cite{pelleg2012can}, is also utilized by people in their search for medical information.

Internet data has been used to study health and medicine, both physical \cite{soldaini2017inferring,white2017evaluation} and mental \cite{de2014mental,stevie2016quantifying}. Specifically, the posts and queries of people who  searched for medical information have been demonstrated to by useful for identifying the illnesses of people before they are aware of them \cite{soldaini2017inferring,white2017evaluation,de2013predicting,hochberg2019can,allerhand2018detecting}. Social media data were used for similar purposes: Reddit data were shown to predict which individuals would transition from mental health discourse to suicidal ideation \cite{de2016discovering}, Twitter posts were utilized to predict future postpartum depression of new mothers \cite{de2013predictingpostpartum} and to estimate the risk of depression \cite{de2013predicting}, and posts from Yahoo Answers to predict the risk of autism from the description of children's behavior \cite{ben2016online}. 

Screening questionnaires are used in medicine as a diagnostic aid. There exist screening questionnaires for several medical conditions, e.g. depression \cite{kroenke2002phq}. These questionnaires are produced on a case-by-case basis, through a long and expensive process \cite{national2014developing}.

Since people describe their experiences on social media, we hypothesize that it may be possible to use the reported symptoms and experiences of people, made prior to a medical diagnosis, in order to construct medically-valid diagnostic questionnaires which could be answered by laypeople to facilitate in their diagnosis. We note that such a questionnaire will differ from one created by medical specialists because people often report symptoms which are different from those that medical practitioners focus on when diagnosing a medical condition \cite{soldaini2016enhancing,ben2016online}. 

Here we focus on diseases where the time between the appearance of the first symptom and the time of diagnosis is long. Thus, earlier detection of these conditions would have a large impact on patient quality of life. Moreover, the long duration allows patients to report (and algorithms detect) symptoms related to the disease in postings made by these users, even when they are not directly attributed to the disease by the users.

Our source of information in this study is Reddit, a popular social networking site that enables users to share text and media and to comment on other people's posts through topic-specific forums. Each forum (colloquially known as a "subreddit") represents a community dedicated to specific interests. Subreddits span a gamut of topics including politics, health, and leisure. Reddit users are not required to identify themselves, and hence many users choose to remain anonymous. Medically-related posts on Reddit are made on disease-specific subreddits (for example, for patients with endometriosis) and on general subreddits, where people often request diagnostic information \cite{shatz2017fast}.


\section{Methods}

The process of building the questionnaire requires 3 stages: Identifying a cohort of users with the disease of interest, finding relevant prior posts, defining a comparable control group, and building a set of decision rules to differentiate between the controls and ill users according to their (prior) posts. 

The first stage of the process is to identify Reddit users that have the condition of interest (the condition cohort). This is achieved by finding  posts in subreddits dedicated to the condition, where the writer mentions that he or she was diagnosed with the condition. 

We trained a classifier to identify such posts, separately for each condition. First, we created a dataset by manually labeling posts for whether the user indicated they had the condition of interest. We then generated a decision tree where the attributes were (1) the bag of words of the text; (2) the existence of external links in the post; and (3) the user's interaction in disease-related subreddits (number of posts, comments, replies, and words). The performance of the model was estimated using leave-one-out cross-validation, and applied to all posts from each disease subreddit. The result of this stage are a list of users that have the condition (e.g., the condition cohort).


The next step is to select the subreddits containing posts from which the questionnaire will be constructed. This is done by collecting posts in all subreddits unrelated to the condition made by people in the condition cohort. Only unrelated subreddits are used here so as to reduce the chance that posts will contain information from the medical inquiry related to the condition. The postings were further filtered to keep those posts written prior to the first post of the user in disease-related subreddits and those posts that contained 80 words or more. After the filtration process, the 13 subreddits that had postings from the most users were selected. 

Subreddits were further filtered so as to focus on subreddits that contained posts that are possibly related to a medical condition: From each of the selected subreddits, 10 posts were sampled randomly and the question "If you had the problem described in the post, would you consult a doctor?” was answered manually. If the answer to at least one post was positive, the subreddit was used in building the questionnaire. The result of this stage are a list of prior posting by users in the condition cohort.

The next step in creating the questionnaire was selecting a control group which reported symptoms similar to those of the condition cohort, but did not write in the condition-related subreddits. First, we found all mentions of the symptoms and their synonyms from \cite{yom2013postmarket}, as well as a list of symptoms constructed during the creation of the cohort identification stage. The most common of these symptoms were used to construct the group of control users, which is comprised of users who mentioned one of these symptoms in the general medical subreddit '/r/AskDocs'. All users included in this group wrote at least 2 posts on Reddit and the total number of words in subreddits relevant to the questionnaire was at least 80.

The final questionnaire was created by training a number of decision trees with a depth of 6, that separated the condition cohort from the controls at the user-level. The depth was chosen so as not to create overly long sequences of questions in the questionnaire. The difference between the trees was their features, created as follows: Word Mover's\cite{kusner2015word} was used to estimate the distance between symptoms in the above-mentioned list. The symptoms were then clustered using agglomerative clustering, in order to group symptoms with similar meaning. We varied the number of clusters between 5 and the total number of symptoms. For each cluster size, a decision tree was created, such that its attributes are whether a user mentioned one of the symptoms in the cluster or not. The performance of the classifier was estimated using leave-one-out cross-validation. We selected the tree from which to create a questionnaire by finding the tree that achieved high AUC \cite{duda2012pattern} and where the clustering did not group too many symptoms into any single cluster, so as to make them unusable.


The final step in creating the questionnaire was to transform the decision tree into a list of question groups. Each path in the tree was converted into a series of questions. To create each question we extracted 3 posts containing each of the symptoms in a cluster (corresponding to a tree node). These posts were read by one of the authors, and a logical sentence containing the symptoms was constructed. For example, the cluster that contains the symptoms "lose weight" and "weight loss" was translated into the question: 'Did the patient mention that he\textbackslash she lost weight?'. 

To assess the validity of the questionnaire we recruited the assistance of medical doctors and asked them to score each path in the decision tree, instructing them: "In your opinion, would a patient that responded to the questions as shown should see a doctor to be screened for the possibility that he\textbackslash she is suffering from the condition?", on a five-point scale. Additionally, doctors had the option to choose “Not enough information”. Pearson’s correlation was calculated between doctors' score and node probabilities of the decision tree. Reliability of the doctors was assessed by showing 50\% of the paths twice to each doctor and calculating the correlation between the scores given by the doctors to the same paths. 

\section{RESULTS}

\begin{table}[tb]
\caption{Data used to build the questionnaire as well as the AUC of the classifier for identifying the postings where users indicate their condition.}
\begin{small}
\begin{tabular}{lccccc}
\toprule
Condition       & Condition & Control & \# of    & \# of    & AUC  \\
              & cohort    & group   & clusters & symptoms &      \\
                \midrule
Endometriosis & 172       & 273     & 75       & 120      & 0.74 \\
Lupus         & 117       & 344     & 55       & 93       & 0.68 \\
Gout          & 154       & 247     & 104      & 104      & 0.71\\
  \bottomrule
\end{tabular}
\end{small}
\label{table:Table1}
\end{table}

\begin{table}[tb]
\caption{Questionnaire AUC and the average correlation between medical doctor's scores and questionnaire scores.}
\begin{tabular}{lcc}
\toprule
Condition       & AUC           & Avg.  correlation      \\
              & questionnaire & med. doctor   \\
              &               & and questionnaire \\
 \midrule
Endometriosis & 0.60          & 0.58                  \\
Lupus         & 0.61          & 0.40                   \\
Gout          & 0.68          & 0.27                 \\
\bottomrule
\end{tabular}
\label{table:Table2}
\end{table}

The process described in the Methods was applied to three conditions that are difficult to diagnose and have a major impact on patients' lives: Endometriosis, Lupus and Gout  \cite{kuznetsov2017diagnosis,thong2017systemic,zhang2006eular}. Table \ref{table:Table1} shows the number of people in the condition cohort and in the control group, number of clusters in the chosen decision tree, the number of symptoms in the decision trees, and the AUC of the classifier which identifies the condition cohort from posts in the condition subreddit. We note that average AUC across conditions was 0.71 (when the  classifier was trained for and applied to the same condition). Applying the model for one condition to the other degraded the average AUC to 0.59, indicating that it is preferable to train a separate classifier for each condition. The average consistency among doctors, as measured by the correlation of scores given to the same items at different times was 0.7 for Endometriosis, 0.95 for Lupus and 0.66 for Gout. 

In the following, we provide details on the construction of the endometriosis questionnaire. The two other conditions underwent a similar process. Endometriosis is a chronic gynecological disorder that affects an estimated 10\% of women of reproductive age \cite{kuznetsov2017diagnosis}. Its symptoms are pelvic pain, painful menstrual periods, and subfertility \cite{kuznetsov2017diagnosis}. Moreover, there is evidence that women with endometriosis are at risk of developing comorbidities such as depression and anxiety disorder \cite{hudelist2012diagnostic}. The time between its first symptom and a definite diagnosis is estimated at between 4 and 10 years \cite{kuznetsov2017diagnosis}. 

There are two dedicated subreddits for women with the disease: ‘/r/endometriosis’ and ‘/r/Endo’. A total of 2,336 users wrote posts in these subreddits. To identify users suffering from endometriosis, the postings of 200 users were manually labeled. Labeler agreement was calculated by randomly selecting 47 users (of the 200 users), which were independently labeled by an additional labeler. Cohen's kappa for the agreement between the two labelers was 0.67, which represents substantial agreement \cite{viera2005understanding}. Using the 200 labeled users as training data, a decision tree model was trained to classify the remaining users. The AUC of this model was 0.74 (see Table \ref{table:Table1}). At the chosen threshold of 50\% (true positive: 0.8, false positive: 0.3), 1,368 people were labeled as ill and 768 as healthy.
 
\begin{figure}[tb]
  \centering
  \includegraphics[width=0.75\linewidth]{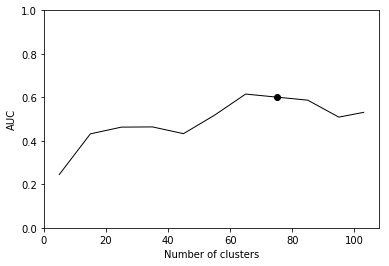}
  \caption{AUC as a function of the number of clusters for endometriosis. The dot marks the selected operating point for generating the questionnaire.}
  \Description{Number of clusters VS AUC Endometriosis.}
  \label{figure:endo_number_cluster_vs_auc}
\end{figure}

\begin{figure*}[h]
  \centering
  \includegraphics[width=0.75\textwidth]{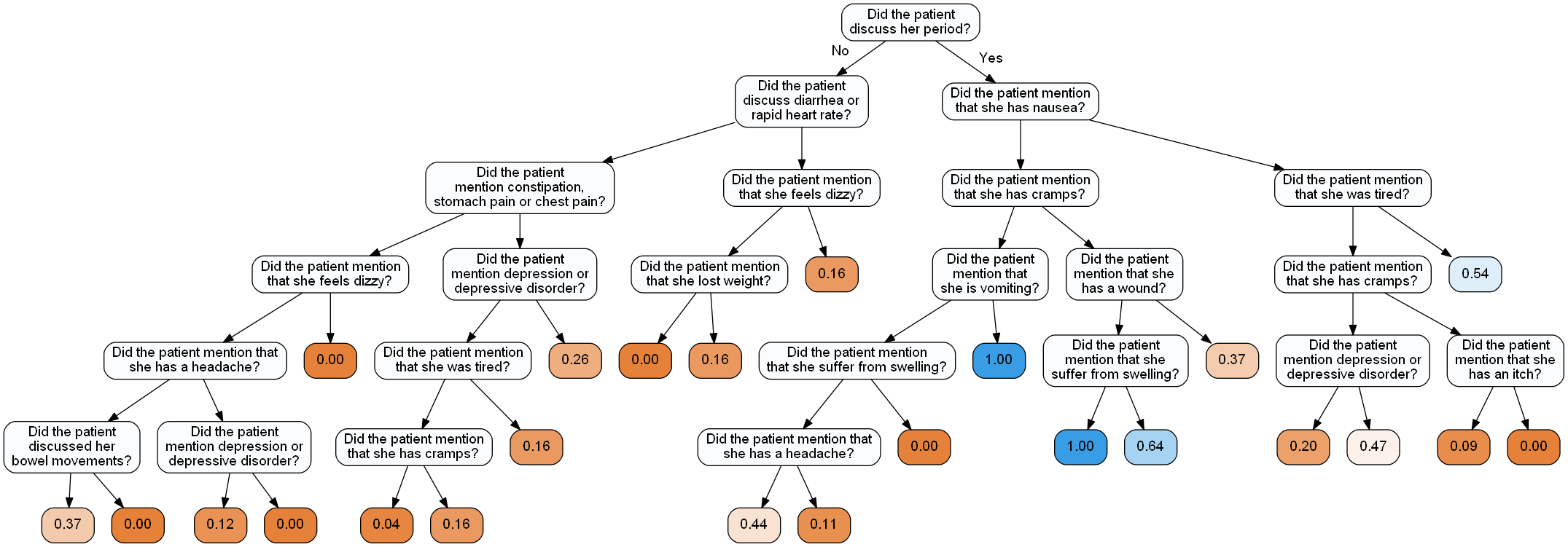}
  \caption{Endometriosis questionnaire. Terminal nodes show the probability that a user, responding to the questions as shown in the path leading to the node, is suffering from endometriosis.}
  \label{figure:endo_Endometriosis_questionnaire}
\end{figure*}

The subreddits from which the questionnaire was constructed were: '/r/TwoXChromosomes', '/r/AskDocs', '/r/AskReddit, \\'/r/depression', '/r/birthcontrol', '/r/Anxiety', '/r/childfree', \\'/r/offmychest', '/r/legaladvice', '/r/sex'. 
Users who wrote in the subreddit '/r/AskDocs' and mentioned ‘pain’, ‘anxiety’ or ‘bloating’ were selected for the control group. Finally, the condition cohort comprised of 172 users and the control group 273 users. In the postings of these users, 120 symptoms were mentioned in the postings. Figure \ref{figure:endo_number_cluster_vs_auc} shows the AUC as a function of the number of clusters. The final questionnaire, shown in Figure \ref{figure:endo_Endometriosis_questionnaire}, has 75 clusters. 
	
The performance of the resulting models is shown in Table \ref{table:Table2}. We note that adding bag-of-word attributes can improve the AUC for endometriosis (0.74) and gout (0.84). However, we chose not to add these attributes because a pre-test showed that they are difficult for patients and medical doctors to interpret and use.

Doctors in the study reported several reservations regarding the questionnaires, which serve as useful information for their further development. First, in some instances they reported that symptoms were grouped for non-obvious reasons (e.g., "stomach pain" and "chest pain"). Second, some paths were too short to allow for reasonable confidence in diagnosis. Finally, the mere mention of several of the symptoms is insufficient information and would additionally require input such as duration, severity, and when the symptom occurs (e.g., pain which appears when exercising). 

\section{DISCUSSION}
Medical questionnaires are diagnostic tools whose production requires considerable time and resources. In this preliminary work, we used social media data from Reddit to (almost fully) automate the production of questionnaires. We applied the process to 3 medical conditions and found good agreement with the scores medical doctors gave to hypothetical examples based on the questionnaires.

Although our questionnaires showed good correlation with medical experts, we assume that the correlation does not show their full diagnostic potential, for several reasons: First, there is a difference in the information that patients share with their physician, as compared to the information they provide on anonymous social media. These differences are due to several reasons, including the variation in reporting frequency: social media is continuously available, compared to (relatively) rare short doctor visits; specificity: doctor visits are very problem-oriented and time-constrained; and testing, which is available to the doctor, but not to people on social media. As a result, the questionnaires sometimes contain risk factors that studies have shown to be linked to the condition, though they are not often taken into account when doctors conduct differential diagnosis. For example, the gout questionnaire included a number of factors that are related to gout, including weight loss \cite{barskova2004insulin,pi1993short}), meat consumption\cite{choi2005gout}, anxiety\cite{prior2014fri0192} and beer consumption \cite{neogi2014alcohol}. Despite this literature, they had the lowest correlation with the medical doctors' scores. When analyzed, doctor’s comments indicated  that some of these factors were dismissed because they were not specific to the condition. For example, a question regarding beer consumption received the following comment: "The majority of the population drink beer". These comments also exemplify the need to include context, duration and timing of symptoms, which the current questionnaires do not allow. For example, if the questionnaire would have asked specifically about the appearance of symptoms after beer consumption, the answer would have been easier to interpret by the physician.

In addition to the above-mentioned limitations of the generated questionnaire, we note two additional drawbacks which need to be addressed in future work: First, the cohort identified in Reddit data might not faithfully represent the demographics of a disease. This is both because of differences in representation of Reddit users compared to the general patient population and because of differences in people's likelihood to declare their illness \cite{yom2019demographic}. Second, the proposed questionnaires were evaluated for their overall accuracy, but not for their precision and recall, compared to the general population. We are working to perform these estimates.

\bibliographystyle{ACM-Reference-Format}
\bibliography{sample-base}

\end{document}